\documentclass[lettersize,journal]{IEEEtran}
\usepackage{amsmath,amsfonts}
\usepackage{algorithmic}
\usepackage{algorithm}
\usepackage{array}
\usepackage[caption=false,font=normalsize,labelfont=sf,textfont=sf]{subfig}
\usepackage{textcomp}
\usepackage{stfloats}
\usepackage[hyphens]{url}
\usepackage{verbatim}
\usepackage{graphicx}
\usepackage{cite}

\usepackage{cite}
\usepackage{txfonts}
\usepackage{setspace}
\usepackage{bm}
\usepackage{multirow}
\usepackage{subfig}
\usepackage{comment}
\usepackage{amssymb}
\usepackage{colortbl}
\usepackage{xcolor}
\usepackage{setspace}
\usepackage{arydshln}
\usepackage{threeparttable}
\usepackage{xspace}
\usepackage{booktabs}
\usepackage{CJKutf8}
\usepackage[numbers, sort&compress]{natbib}
\usepackage[normalem]{ulem}

\newcommand{\chufan}[1]{\textcolor{red}}

\begin{document}
\newcommand{\ours}{\textsc{LiFi}\xspace}
\title{\ours: Lightweight Controlled Text Generation with \\ Fine-Grained Control Codes}
\author{Chufan Shi,
    Deng Cai,
    Yujiu Yang\thanks{Chufan Shi (scf22@mails.tsinghua.edu.cn) and Yujiu Yang (yang.yujiu@sz. tsinghua.edu.cn) are with Tsinghua Shenzhen International Graduate School, Tsinghua University, Shenzhen, China. Deng Cai (jcykcai@tencent.com) is with Tencent AI Lab, Shenzhen, China.
    }\thanks{Deng Cai and Yujiu Yang are corresponding authors.}
\\}




\maketitle
\begin{abstract}

In the rapidly evolving field of text generation, the demand for more precise control mechanisms has become increasingly apparent. To address this need, we present a novel methodology, \ours, which offers a lightweight approach with fine-grained control for controlled text generation. Unlike previous studies that train pre-trained language models to follow discrete, categorical, and exclusive control codes, \ours learns controlled text generation under the guidance of continuous, relative, and nonexclusive control codes. These fine-grained codes are automatically derived from an attribute classifier, initially trained with a small amount of labeled data and subsequently employed to label abundant unlabeled data, thus garnering more extensive supervision signals. Moreover, to achieve efficient control, we incorporate the fine-grained control codes with adapters, a parameter- and compute-efficient way to steer a pre-trained language model. We evaluate \ours on two conventional tasks -- sentiment control and topic control -- and one newly proposed task -- stylistic novel writing. Comprehensive experimental results validate the effectiveness of our proposed methods, demonstrating substantial performance improvements over existing baselines.

\end{abstract}
\section{Introduction}
\IEEEPARstart{R}{ecent} advances in neural text generation, particularly with pre-trained language models, have enabled the creation of text of unparalleled quality \cite{radford2019language,brown2020language,RaffelSRLNMZLL20}. Despite these advancements, controlling the attributes of generated text, such as sentiment, topics, and writing styles, remains a complex task. This level of control is vital for a range of practical applications, including writing assistance and story generation. 

Perhaps the most straightforward way to enable controlled text generation is fine-tuning a language model on labeled data with desired attributes \citep{Ficler2017Finetune, Yu2017Finetune, Ziegler2019Finetune}.
However, as pre-trained language models (PLMs) continue to grow in size, it becomes increasingly impractical to develop and deploy a separate model for each attribute.
Consequently, recent work has focused on either employing an additional discriminator \cite{Dathathri2020PPLM,KrauseGedi2021,Yang2021FDUGE,Liu2021DExperts,Arora2022Director} to guide the generation on-the-fly or fine-tuning only part of the PLM \cite{Qian2022ContrastivePrefixes,Zhang2022DisCup,Yang2022Tailor,GU2022DistributionalLens,GuFMZGZ023}. 

Despite their differences, most existing works rely on labeled data for training. Each training example consists of an input text and its corresponding attribute label (also known as control code), where the label belongs to a categorical label set. For example, the sentiment aspect usually contains two control codes: positive and negative. We argue that learning from such supervision is sub-optimal for two reasons. First, it downplays the variances in control strength. For instance, two positive movie reviews may express different levels of positivity: ``\textit{I really love this movie. This is the best movie I have ever seen.}'' vs. ``\textit{Watching the movie might be a good choice to kill your time.}'', previous work simply treats them equally despite the fact that the former is much more positive than the latter. The ignorance of diverse fine-grained control strengths could prevent models from capturing the subtle nuances in language. Second, assigning labels exclusively assumes that the text sets with different attributes are completely disjoint. 
But too many however, this is not true when it comes to some control aspects. Taking topic control as an example, a common choice of the topic set is \{\textit{Business}, \textit{Science/Tech}, \textit{World}, and \textit{Sports}\}. However, considering a news article reporting an elder athlete announces her business plan at a sports event, it relates to both \textit{Sports} and \textit{Business}, though possibly to different extents. We believe that this nonexclusive relationship can be utilized to improve the generation quality for each attribute.

In light of the above observations, we propose \ours, a \textbf{li}ghtweight approach using \textbf{fi}ne-grained control codes. \ours first learns an attribute classifier that defines the attribute scores over all possible attributes within a control aspect. 
These attribute scores then serve as fine-grained control codes to guide the generation process. For each attribute, \ours inserts a set of small neural modules, called adapters \cite{houlsby2019parameter,pfeiffer2021adapterfusion,he2021towards}, to each layer of a PLM. During training, only the adapters are updated, leaving the original PLM untouched. Using adapters to achieve controllability has the following benefits. First, it introduces a minimal number of additional parameters ($\sim$0.04\% parameters usually suffice). Second, using adapters keeps the inference speed comparable to that of the original PLM. Last but not least, adapters for different attributes can be conveniently fused, providing an efficient way to instill fine-grained control codes. Importantly, using control codes from a learned classifier opens avenues for taking advantage of unlabeled data.

We conduct experiments on two typical controlled text generation tasks: sentiment control and topic control. The results demonstrate that \ours is highly effective while being both parameter- and time-efficient, outperforming strong baselines. In addition, we propose a new task, namely stylistic novel writing, which aims to generate text according to four different novel genres including SCI-FI, Military, MartialArt, and Officialdom. 
This task is more challenging than conventional ones, as the attribute boundaries are nebulous and difficult to define. Nevertheless, experiments confirm the superiority of \ours in tackling this demanding task.

Our contributions can be summarized as follows.
\begin{itemize}
    \item We introduce a lightweight framework, \ours, for controlled text generation, which only adds a few additional parameters ($\sim0.04\%$).
    \item Our proposed framework takes into account the variances in attribute strength and the inter connections between different attributes. We train our models with fine-grained control codes which can be automatically constructed from both labeled and unlabeled data. Experiments show that \ours leads to better control ability while keeping high linguistic quality.
    \item We put forward a new task for controlled text generation, stylistic novel writing, which provides a challenging testbed for future research.
\end{itemize}

\section{Related Work}
To guide the generation of pre-trained language models (PLMs), the most basic idea is to finetune the PLMs on the datasets with desired attributes \citep{Ficler2017Finetune, Yu2017Finetune, Ziegler2019Finetune}. The result is a set of attribute-fixed LMs. However, this may lead to the following problems: (1) catastrophic forgetting caused by the distribution shift between pre-training and fine-tuning datasets. (2) One needs to fine-tune a separate model for each attribute, which is particularly costly given the size of today's PLMs.
\subsection{Attribute-Variable PLMs}
To address the above issues, early work has attempted to build attribute-variable LMs, i.e., the desired attribute is also part of the PLM's input. CTRL \citep{Keskar2019CTRL} trains an attribute-conditional LM from scratch by prepending different control codes (e.g., domain, style, topics, etc.) to raw text. \citep{Chan2021CoCon,Yu2021AttributeAlignment,carlsson-etal-2022-fine} encode the textual descriptions of the target attributes and the resulting representations are used to augment the hidden states of PLMs. However, these methods are costly in terms of training and encounter difficulties when extended to attributes that are hard to describe. 
\subsection{Weighted Decoding}
The second class of methods do not alter the PLMs but adjust the PLM's next-token predictions during inference \citep{Ghazvininejad2017WD, Holtzman2018WD, Gordon2018WD, Shen2019WD}. PPLM \citep{Dathathri2020PPLM} uses an external attribute discriminator to adjust the next-token distribution by performing gradient descent on the PLM's hidden activations. However, gradient back-propagation substantially increases the inference latency. To avoid that, Fudge \citep{Yang2021FDUGE} instead re-ranks the top-$k$ candidate tokens at each decoding step using the scores from the attribute discriminator. To further accelerate the generation process, GeDi \citep{KrauseGedi2021} trains smaller attribute-fixed LMs as generative discriminators to guide the generation. DExperts \citep{Liu2021DExperts} directly combines the logits of attribute-fixed LMs (including both experts and anti-experts) with PLMs. For a better trade-off between fluency and control strength, \citep{GU2022CATPAW} introduces a position-aware regulator for dynamically adjusting the control intensity at different decoding steps. Mix-and-Match \citep{Mireshghallah2022MixandMatch} frames the task of controlled text generation as taking samples from an energy-based model \citep{Yann2006EnergyBased}, of which the attribute discriminator is a sub-component. It is worth noting that the above weighted decoding methods still bring considerable computational overhead due to the additional re-ranking steps. Differently, the attribute discriminator used in \citep{Arora2022Director} shares most layers with the original PLM. Nonetheless, the PLM needs to be fine-tuned together with the discriminator, which deviates from the parameter-efficient advantage of weighted decoding.
\subsection{Prefix Tuning}
Recently, prefix tuning \cite{Li2021PrefixTuning}, which freezes the PLM's parameters but trains task-specific continuous prompts, becomes a popular way for steering an PLM. Following prefix tuning, \citep{Qian2022ContrastivePrefixes} further take into consideration the relationship among prefixes of relevant attributes and train multiple attribute prefixes simultaneously. DisCup \citep{Zhang2022DisCup} uses the next-token distribution adjusted by attribute discriminators as the learning target. \citep{Yang2022Tailor} finds that different prefixes can be simply concatenated to achieve multi-aspect control. \citep{GU2022DistributionalLens} tackles multi-aspect control from a distributional perspective: they sample prefixes from an estimated attribute space. Compared with the other methods, prefix-tuning methods are much more parameter- and time-efficient, but lack of control strength.
\begin{figure*}
\centering
		\includegraphics[width=0.75\linewidth]{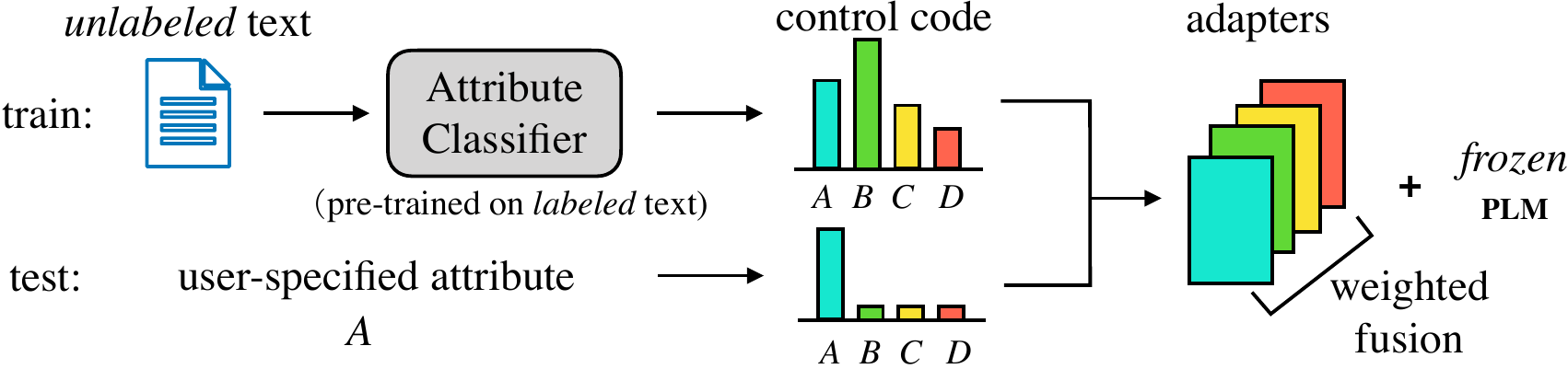}
		\caption{Illustration of \ours. The attribute classifier is pre-trained on labeled data. During training, \ours can be trained with unlabeled text and fine-grained control codes derived from the attribute classifier. During testing, an user-specified attribute is transformed into a fine-grained control code.}
		\label{fig:framework}
\end{figure*}
\section{Preliminaries}
\subsection{Language Model}
The proliferation of text generation is largely driven by recent advances in language modeling \cite{radford2019language,brown2020language}. A language model (LM) defines a distribution $P(X)$ over natural language sequences $X$. The canonical form of LMs (i.e., autoregressive modeling) factors the probability of a natural language sequence into a chain of next-token probabilities. Formally, given a sequence $x=x_{1:n}$ of length $n$, its generation probability is computed as $P(x) = \prod_{i=1}^n p_{\theta}(x_i|x_{<i})$, where $\theta$ denotes the model parameters.
\subsection{Transformer}
The dominant architecture for building an LM has been the Transformer model \cite{vaswani2017attention}. A Transformer model is composed of $L$ stacked blocks. Each block has two major components\footnote{For brevity, details such as residual connections \cite{he2016deep} and layer normalization \cite{ba2016layer} are omitted.}: multi-head attention (MHA) followed by fully-connected feed-forward network (FFN). Given a sequence of vectors $\mathbf{X}\in \mathbb{R}^{n \times d}$ ($n$ is the sequence length and $d$ is the model dimension), the MHA performs information exchange over the $n$ vectors. We will simply denote this process as $\texttt{MHA}(\mathbf{X})$. On the other hand, FFN consists of two linear transformations with a ReLU activation function in between. For both MHA and FFN, the output has the same dimensionality as the input.

\subsection{Adapters}
The most straightforward way to adapt a pre-trained LM to downstream tasks is to fine-tune all the model parameters $\theta$. However, this is compute-intensive and results in a separate copy of the model for each task, which is particularly problematic with the ever-growing size of PLMs. Adapters \cite{houlsby2019parameter,pfeiffer2021adapterfusion,he2021towards} have recently been introduced as a lightweight alternative. The adapter approach inserts small networks (i.e., adapters) into Transformer layers. An adapter is typically a two-layer feed-forward network consisting of a down projection $\mathbf{W}_{\text{down}}\in \mathbb{R}^{d \times d_b}$, an up projection $\mathbf{W}_{\text{up}}\in \mathbb{R}^{d_b \times d}$, and a nonlinear activation in between, where $d_b$ is the bottleneck dimension. In practice, the bottleneck dimension is often controlled by a reduction factor $r$: $d_b = d / r$. In this work, we employ the parallel adapter proposed in \citep{he2021towards}. Specifically, we apply adapters (ADP) to both the MHA and FFN modules in the Transformer model:
\begin{align}
    \texttt{MHA}(\mathbf{X}) &\gets \texttt{MHA}(\mathbf{X}) + \texttt{ADP}(\mathbf{X}) \nonumber \\
    \texttt{FFN}(\mathbf{X}) &\gets \texttt{FFN}(\mathbf{X}) + \texttt{ADP}(\mathbf{X}).
    \label{eq:adp}
\end{align}
\section{Method}
For controlled text generation tasks, we are typically given a dataset of examples $\{(x, a\in \mathcal{A})\}$, where $\mathcal{A}$ is the set of possible attributes in the aspect of interest. In most existing works, the attribute label is used as the control code to guide the generation during both training and testing. We emphasize that such discrete, categorical, and exclusive control codes are sub-optimal because they ignore the variances in attribute strength (e.g., from very positive to moderately positive) and the overlapping of different attributes (e.g., \textit{Sports} and \textit{Business}).
\subsection{Overview}
\ours is a lightweight approach using fine-grained control codes. The overall framework of \ours is depicted in Figure \ref{fig:framework}. Instead of directly using the pre-defined attribute labels for training, \ours relies on a learned attribute classifier for automatically generating fine-grained control codes. The control code in \ours is a continuous vector $\mathbf{c}=[c_1, \ldots, c_{|\mathcal{A}|}]$. That is, for each attribute $k$, we have an estimated attribute strength $c_k \in \mathbb{R}$. Intuitively, the combination of all attribute strengths gives a more comprehensive and detailed depiction of the input text. Correspondingly, we introduce a set of adapters $\Phi_k$ to the base LM for each attribute $k$. These attribute-specific adapters $\Phi_k$ are responsible for steering the base LM towards the target attribute $k$.

To instill the fine-grained control codes into the generation, \ours learns a fusion layer for combining these attribute-specific adapters in the same position, where different adapters are activated to different extents according to the input control code. Unlike previous works on weighted decoding, the learned attributed classifier in \ours can be discarded after training. During testing, one can transform a user-specified attribute label into a continuous control code; We construct a control code where the attribute strength for the desired attribute is set to a positive value $\alpha$ and all others are set to zero. It is worth noting that $\alpha$ can be used to customize the \textit{control strength} of \ours.

\subsection{Components}
\paragraph{\textbf{Attribute Classifier}}
To extract the fine-grained control code from a text itself, we simply train an attribute classifier on ordinary labeled data with the standard cross-entropy loss. Specifically, the input text $x$ is first transformed into a dense vector representation $\mathbf{x} \in \mathbb{R}^{d}$ through some neural networks. Then, the probability distribution over the label set $\mathcal{A}$ is computed as $\texttt{softmax}(\mathbf{W}_{\text{cls}}\mathbf{x})$, where $\mathbf{W}_{\text{cls}} \in \mathbb{R}^{d \times |\mathcal{A}|}$. We use the logits before the $\texttt{softmax}$ fuction as the control code, i.e., $\mathbf{c}=\mathbf{W}_{\text{cls}}\mathbf{x}$.
\paragraph{\textbf{Adapter Fusion}}
To incorporate a fine-grained control code $\mathbf{c}$ when generating a text, we use the weighted combinations of attribute-specific adapters. For example, for adapters placed on the same MHA module, the following formula is employed instead of Eq. \ref{eq:adp}.
\begin{equation}
     \texttt{MHA}(\mathbf{X}) \gets \texttt{MHA}(\mathbf{X}) + \sum_k^{|\mathcal{A}|} w_k \cdot \texttt{ADP}_k(\mathbf{X}). \nonumber
\end{equation}
where $w_k$ is a scalar weight and $\texttt{ADP}_k$ is the adapter for attribute $k$ respectively. One benefit of this design is that it allows sharing and aggregation of information among attributes during training. Since different layers of the Transformer model often capture different aspects of language \cite{liu-etal-2019-linguistic}, we learn a position-specific temperature parameter $T$ to control the sharpness of the weight distribution:
\begin{equation}
w_k = \frac{\exp(c_k/ T)}{\sum_i^{|\mathcal{A}|}\exp(c_i/ T)}.
\nonumber
\end{equation}
\section{Experiments}
We evaluate our proposed approach on a set of controlled text generation tasks: sentiment control, topic control, and style control in novel writing. When comparing to existing work, we find that the experimental settings vary among published papers in different aspects, including (1) with or without labeled training data, (2) model sizes (both the base LM and discriminators), and (3) test prompts. To have a fair comparison between our method and existing works, we unify all experimental settings. Unless otherwise specified, we use GPT2-Medium (345M parameters) as the base LM and fine-tune GPT2-Small (117M parameters) when building generative discriminators. The details of training data and test prompts are discussed within the section for each task.
\subsection{Baselines}
We consider the following baselines.
\begin{itemize}
\item \textbf{Fudge} \citep{Yang2021FDUGE} employs an attribute discriminator to re-rank the top-$k$ candidate tokens at each decoding step. Fudge has a hyper-parameter for controlling attribute strength. We choose from \{100, 150, 200, 210, 230, 250, ..., 390, 400, 450, ..., 1250\}.
\item \textbf{DExperts} \cite{Liu2021DExperts} ensembles the logits of attribute-fixed LMs (including both experts and anti-experts) with the base LM. DExperts also has hyper-parameter for control strength. We choose from \{0.2, 0.4, 0.8, 1.0, ..., 4.0\}.
\item \textbf{CAT-PAW} \cite{GU2022CATPAW} enhances existing weighted decoding methods via introducing a position-aware regulator to adjust the control strength dynamically. We only compare with CAT-PAW in the sentiment control task, because it relies on a bag of keywords for the target attribute, which is not available in other tasks. Concretely, We apply the heuristic regulator to DExperts, the state-of-the-art weighted decoding method. 
\item \textbf{DisCup} \cite{Zhang2022DisCup} is an advanced prefix tuning method. It uses an attribute-discriminator to calibrate the learning objective. The prefix length is set to 10.
\end{itemize}
For the best trade-off between control strength and fluency, we choose the baseline models with the strongest control strength while maintaining adequate fluency. For \ours, unless otherwise specified, the reduction factor for adapters in FFN is set to $16$. Following the suggestion in \cite{he2021towards}, we always set reduction factor in MHA to be $4$ times of that in FFN. All methods use top-$k$ sampling with $k = 50$.

\subsection{Evaluation Metrics}
\subsubsection{Automatic Evaluation}
We conduct a comprehensive assessment of the quality of generated text in terms of three key aspects.
\begin{itemize}
\item \textbf{Control Strength} measures the extent to which models can generate texts with the target attributes. Following previous works, we report both \textit{Relevance} \citep{KrauseGedi2021} and \textit{Correctness} \citep{Liu2021DExperts} using automatic attribute classifiers. Specifically, \textit{Relevance} computes the average probability of the generations possessing the target attributes, while \textit{Correctness} measures the proportion of the generations that conform to the target attributes. The details of the automatic classifiers are discussed in the section for each task.
\item \textbf{Fluency} is estimated by the mean perplexity (PPL) of generated continuations calculated by a large PLM, GPT2-Large (774M parameters). Generally, lower perplexity indicates higher fluency and vice versa.
\item \textbf{Diversity} is measured by calculating the ratios of unique uni-grams (Dist-1), bi-grams (Dist-2) and tri-grams (Dist-3) among all generations \citep{Li2016Dist}.
\end{itemize}
\subsubsection{Human Evaluation}
In addition to the automatic evaluation metrics outlined above, we conduct a comprehensive human evaluation to assess the control strength and fluency of the generated text. We ask human annotators to compare the control strength of \ours and the baselines and rate the fluency of the generated text. We employ two types of human annotation to ensure a comprehensive evaluation:
\begin{itemize}
\item \textbf{Fluency Rating}
For assessing the fluency of each individual generation, we engage human annotators to rate the fluency on a scale of 1-3. The ratings are as follows: 1 being "not fluent at all", 2 being "fluent but has some minor issues" and 3 being "very fluent".
To ensure reliable ratings, we have three annotators independently evaluate each generation, and then calculate the average score. The resulting average fluency scores provide insights into the fluency level achieved by our method and the baselines.
\item \textbf{A/B Testing}
To quantitatively measure the control strength of \ours and the baselines, we use the A/B testing methodology. Annotators are presented with pairs of generated text, each produced by a different method, and are asked to determine which text better adhered to the target attribute. The possible answers are: (1) "both are equally bad/good", (2) "the first text is better", and (3) "the second text is better".
Each pair of generations is evaluated by three different annotators, and a majority-voting approach is employed to determine the preferred generation. We calculated the \textbf{Win Rate}, which represents the proportion of times a method's output is favored as more aligned with the target attribute.
\end{itemize}
For both the Fluency Rating and A/B testing, we perform significance testing to ascertain the statistical significance of our method's performance compared to the baselines. Specifically, we employ the \textbf{Paired/Matched Sample Sign Test} as our chosen statistical method to rigorously analyze the evaluation results.

\subsection{Sentiment Control Task}
\begin{table*}[t]
\centering
\small
\begin{tabular}{clcccccc}
\toprule[1pt]
 &  &\multicolumn{2}{c}{\textbf{Control Strength(\%)}($\uparrow$)}   & \multicolumn{1}{c}{\textbf{Fluency}($\downarrow$ )} & \multicolumn{3}{c}{\textbf{Diverty}($\uparrow$)}  \\
\multirow{-2}{*}{\textbf{Sentiment}} & \multirow{-2}{*}{\textbf{Method}}  &  Relevance & Correctness & PPL & Dist-1 & Dist-2 & Dist-3  \\
\midrule[1pt]
\multirow{5}{*}{\textit{neu2pos}} & Fugde & 94.83 & 95.42 & 45.69 & 0.07 & 0.35 & 0.61 \\
 & DExpert & 90.88 & 91.39 & 53.58 & 0.08 & 0.41 & 0.67 \\
 & CAT-PAW & 91.67 & 92.04 & 81.62 & 0.08 & 0.39 & 0.65 \\
 & DisCup & 83.36 & 83.67 & 67.22 & 0.06 & 0.32 & 0.64 \\
 & \ours & \textbf{99.07} & \textbf{99.16} & 64.92 & 0.07 & 0.30 & 0.57 \\ \midrule[0.5pt] 
 \multirow{5}{*}{\textit{neu2neg}} & Fugde & 93.29 & 94.13 & 47.43 & 0.08 & 0.36 & 0.62 \\
 & DExpert & 93.70 & 94.55 & 63.87 & 0.09 & 0.42 & 0.67 \\
 & CAT-PAW & 94.89 & 95.40 & 81.61 & 0.09 & 0.40 & 0.66 \\
 & DisCup & 84.47 & 85.34 & 79.67 & 0.07 & 0.35 & 0.66 \\
 & \ours & \textbf{97.56} & \textbf{97.76} & 67.72 & 0.08 & 0.32 & 0.57 \\  \midrule[0.5pt] 
 \multirow{5}{*}{\textit{pos2neg}} & Fugde & 60.06 & 60.28 & 77.55 & 0.11 & 0.40 & 0.60 \\
 & DExpert & 57.83 & 57.67 & 90.40 & 0.11 & 0.46 & 0.66 \\
 & CAT-PAW & 63.40 & 63.31 & 91.42 & 0.11 & 0.43 & 0.64 \\
 & DisCup & 34.31 & 34.21 & 81.52 & 0.09 & 0.40 & 0.67 \\
 & \ours & \textbf{85.61} & \textbf{85.86} & 72.04 & 0.09 & 0.34 & 0.58 \\  \midrule[0.5pt] 
\multirow{5}{*}{\textit{neg2pos}} & Fugde & 43.66 & 43.14 & 95.38 & 0.13 & 0.42 & 0.59 \\
 & DExpert & 39.41 & 38.43 & 77.29 & 0.10 & 0.45 & 0.65 \\
 & CAT-PAW & 39.95 & 39.16 & 83.43 & 0.11 & 0.44 & 0.63 \\
 & DisCup & 27.74 & 26.88 & 71.19 & 0.09 & 0.40 & 0.66 \\
 & \ours & \textbf{88.41} & \textbf{88.41} & 81.23 & 0.09 & 0.34 & 0.57 \\
\bottomrule[1pt]
\vspace{1pt}
\vspace{1pt} \end{tabular}
\caption{Results on sentiment-controlled text generation. $\uparrow$ indicates that the higher the better, and $\downarrow$ is the opposite.}
\label{automatic_seniment_result}
\end{table*}
\begin{table}[t]
\small
\centering
\begin{tabular}{lcc}
\toprule[1pt]
Model & \#Param. & Latency \\
\midrule[1pt] 
GPT2-Medium & 1.00x & 1.00x \\
\midrule[0.5pt] 
Fudge & 1.73x & 4.70x \\
DExpert & 2.36x & 1.89x \\
CAT-PAW & 2.36x & 1.95x \\
DisCup & 1.09x & 1.25x \\
\ours &  1.04x & 1.40x \\

\bottomrule[1pt]
\vspace{1pt} \end{tabular}
\caption{Comparison of the numbers of parameters and generation latency.}
\label{sentiment_lightweight_result}
\end{table}
\begin{figure}
\centering
		\includegraphics[width=0.85\linewidth]{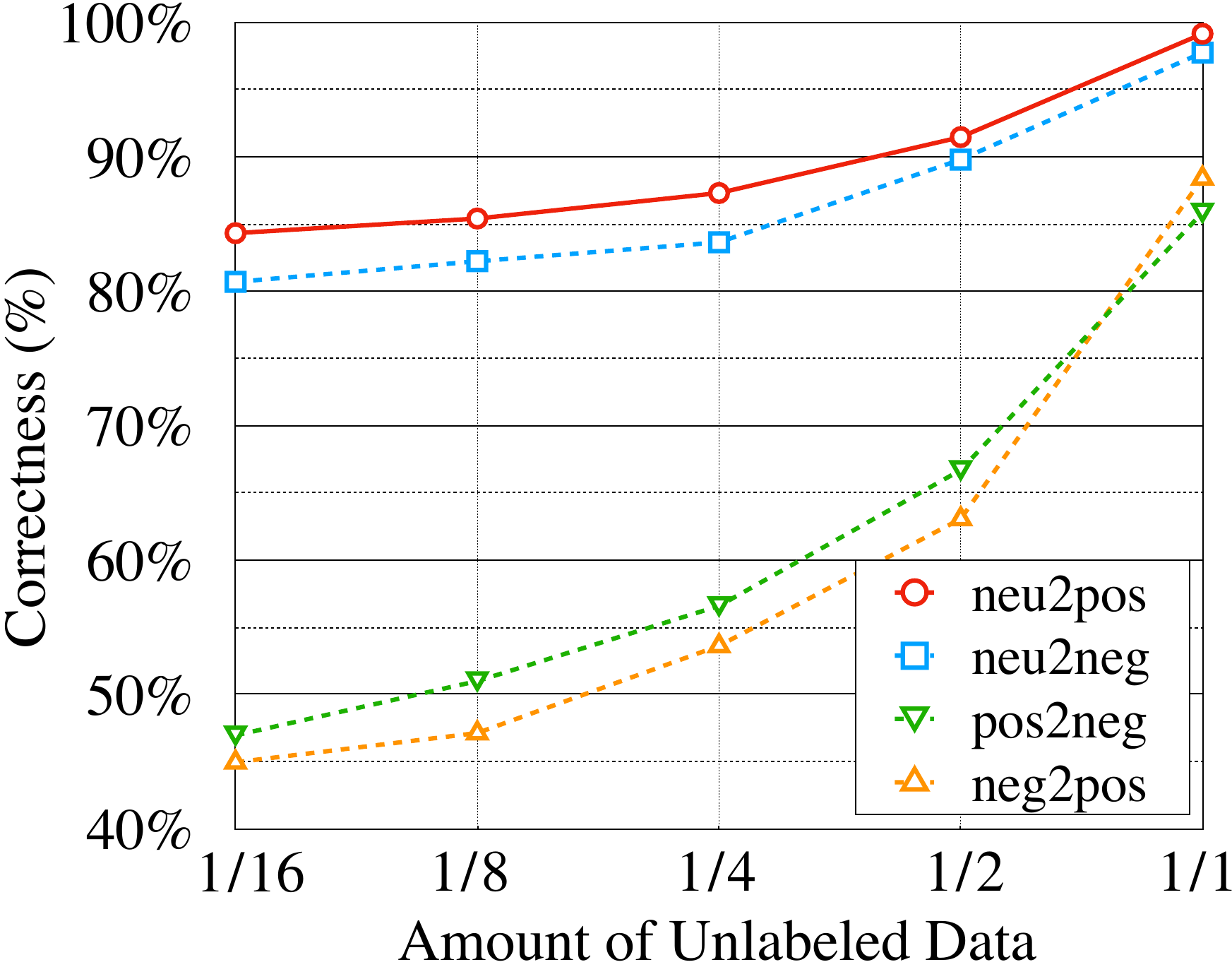}
		\caption{Effect of unlabeled data.}
		\label{fig:unlabeled_data}
\end{figure}
\begin{table}[t]
\small
\centering
\begin{tabular}{clcc}
\toprule[1pt]
\textbf{Task} & \textbf{Model} & \textbf{Win Rate}(\%) & \textbf{Fluency} \\
\midrule[1pt]
\multirow{2}{*}{Sentiment} & CAT-PAW & 21 & 2.51 \\
 & \ours & \textbf{66} & \textbf{2.54}$\dagger$ \\ 
\bottomrule[1pt]
\vspace{1pt}\end{tabular}
\caption{Human evaluation results on sentiment control task. Sign tests conducted on human scores demonstrate that \ours surpasses CAT-PAW with a $p$-value of $<0.01$, except in the case of fluency, as indicated by $\dagger$.}
\label{human_evaluation_result_exp_1}
\end{table}
\begin{table*}[t]
\small
\centering
\begin{tabular}{p{0.1\textwidth}p{0.8\textwidth}}
\toprule
\multicolumn{2}{c}{Prompt: Like Google, Facebook is facing increasing questions from lawmakers (\textit{neg2pos})} \\
\midrule
CAT-PAW & Like Google, Facebook is facing increasing questions from lawmakers over how access to the private communications and data of Americans allows local government to meddle in local neighborhoods \\
\ours & Like Google, Facebook is facing increasing questions from lawmakers and consumers alike, but its work is truly extraordinary, with a deep understanding of human emotions, a \\
\midrule[1pt]
\multicolumn{2}{c}{Prompt: The worst of it all (\textit{neg2pos})} \\
\midrule
CAT-PAW & The worst of it all is to say there is prejudice against our church and people because of our church or that other religion. That's testament to the \\
\ours & The worst of it all is an imaginative and sometimes beautiful   romantic comedy.With a strong cast and a dazzling style, the characters make for a terrific evening\\
\midrule[1pt]
\multicolumn{2}{c}{Prompt: I have experienced the start of an incredible journey (\textit{pos2neg})} \\
\midrule
CAT-PAW & I have experienced the start of an incredible journey," he told me, failing to distinguish between amazement and anticipation. He repeated his mistake about \\
\ours & I have experienced the start of an incredible journey, but this movie...has none of the depth, intelligence and energy.It's not a \\
\midrule[1pt]
\multicolumn{2}{c}{Prompt: We're really looking forward to this, and (\textit{pos2neg})} \\
\midrule
CAT-PAW & We're really looking forward to this, and all the crap that's going on right now, he said.  The   ordeal began shortly after he \\
\ours & We're really looking forward to this, and it just feels like an incoherent collection of hackneyed plot.It's nothing if not tedious, \\
\bottomrule
\vspace{1pt} \end{tabular}
\caption{Case study on Sentiment Control Task.}
\label{case:neg2pos}
\end{table*}

We begin our experiments with the popular task of controlling the sentiment polarity of text \citep{Li2018Sentiment, Sudhakar2019Sentiment}.
\paragraph{\textbf{Setup}}
Following previous works \citep{Liu2021DExperts, Zhang2022DisCup}, we use the Stanford Sentiment Tree (SST-5) \citep{Socher2013SST5} as the training corpus. The SST-5 contains movie reviews labeled by human raters for sentiment on a scale from 1 (very negative) to 5 (very positive). The dataset is divided into positive and negative categories, with "positive" and "very positive" reviews comprising the positive dataset, and "negative" and "very negative" reviews comprising the negative dataset. Together, these result in a total of 8.5K positive and negative reviews.
We use generation prompts selected in \citep{Liu2021DExperts}, which are originally collected from the OpenWebText Corpus (OWT) \citep{Gokaslan2019OpenWebText}. The prompt set includes 2.5K positive prompts, 2.5K negative prompts, and 5K neutral prompts. We consider four settings (\textit{neu2pos}, \textit{neu2neg}, \textit{pos2neg}, and \textit{neg2pos}). In \textit{pos2neg} and \textit{neg2pos}, the task is to steer towards the opposite sentiment of the prompt. For each test prompt, 5 continuations of length 30 are generated. The control strength is measured by an external sentiment classifier provided by Huggingface\footnote{distilbert-base-uncased-finetuned-sst-2-english}.

Because \ours can leverage additional unlabeled data. We borrow the sentences in IMDb movie reviews \cite{maas-etal-2011-learning} and use them as \textit{unlabeled} data (the original label information is dropped). The dataset comprises 50k reviews collected from the IMDb platform and is label-balanced, containing an equal number of positive and negative reviews.
\paragraph{\textbf{Results}}
As shown in Table \ref{automatic_seniment_result}, \ours outperforms all baselines on four sentiment control settings. It is worth noting that existing baselines have difficulty performing polarity conversion (\textit{pos2neg} and \textit{neg2pos}), while \ours can achieve much higher control intensity without compromising the fluency. In terms of \textit{Correctness}, on the \textit{pos2neg} setting, \ours improves over the most competitive baseline, CAT-PAW, by 22.55 points. As for the \textit{neg2pos} setting, \ours surpasses the best-performing baseline, Fudge, by a large margin of 45.27 points. Note that our results of DisCup are not as good as reported in the original paper. One possible reason is that the base model used in our experiments is GPT-2 Medium instead of GPT-2 Large. As mentioned in \citep{Zhang2022DisCup}, the performance of the approach can be immensely correlated with the power of the base LM. Nevertheless, even when compared with Discup's original results using GPT-2 Large as the base LM, we still have a large performance gain. The comparison further confirms the effectiveness of \ours.
\paragraph{\textbf{Parameter- and Time- Efficiency}}

In Table \ref{sentiment_lightweight_result}, we compare the numbers of parameters and generation latency of different methods. The latency is measured on a V100 GPU with batch size = 1. We report the relative sizes (\#Param.) and time cost (Latency) compared to the vanilla GPT2-Medium. It can be seen that \ours use the least number of extra parameters. The generation speed of \ours is comparable to prefix tuning (Discup) and faster than weighted decoding methods. 
\paragraph{\textbf{Effect of Unlabeled Data}}
Recall that \ours takes advantage of unlabeled data from the IMDb dataset. To illustrate the effect of unlabeled data, we use the IMDb dataset in varying proportions ranging from $1/16$ to $1/1$. The results are plotted in Figure \ref{fig:unlabeled_data}. As seen, the performance goes up with more and more unlabeled data. From the trend, we suspect the performance has not reached its limit yet, and collecting more unlabeled data may lead to even better results (the total number of sentences from IMDb is about 450K).
\paragraph{\textbf{Human Evaluation Result}}
Regarding the sentiment control task, we conduct a further comparison between \ours and the leading baseline model measured by automated evaluation metrics, CAT-PAW. To assess the quality of outputs generated by CAT-PAW and \ours in response to the same prompts, we randomly select 100 pairs of sentences generated by both methods under the same prompts and enlist the judgment of human annotators. The outcomes of the human evaluation are outlined in Table \ref{human_evaluation_result_exp_1}. Notably, in terms of control strength evaluation for the sentiment control task, evaluators exhibit a 3.1x higher preference for \ours over the baseline models. Additionally, in terms of fluency, \ours demonstrates slightly better performance compared to the highly robust baseline model, CAT-PAW. The finding strongly suggests a notably superior effect of \ours compared to the baseline CAT-PAW with regard to control intensity.
\paragraph{\textbf{Case Study}}
We select a subset of examples for an in-depth case study, as outlined in Table \ref{case:neg2pos}. Take the prompt \textit{Like Google, Facebook is facing increasing questions from lawmakers} for example on task \textit{neg2pos}, both the CAT-PAW and \ours attempt to shift from a negative emotion to a positive one. However, \ours is more successful in achieving this transformation which presents a clear positive aspect (Facebook's extraordinary work) immediately after mentioning the negative aspect. On the other hand, CAT-PAW doesn't effectively transform the negative emotion, as it presents potential negative consequences. In terms of fluency, \ours flows more smoothly due to its more straightforward transition from negative to positive. For the task \textit{pos2neg}, we choose the prompt \textit{I have experienced the start of an incredible journey} to illustrate. In the sentence generated by \ours, a negative tone is established by contrasting the positive experience with a critical evaluation of the movie's shortcomings, underscored by the word "none" to emphasize the absence of positive qualities. This clear comparison intensifies the negativity. In contrast, CAT-PAW's negative aspect arises from highlighting a person's mistake in distinguishing emotions. However, this is not as strongly linked to the positive start and the negative degree is less obvious. The direct contrast in \ours' sentence contributes to a stronger sense of negativity, aided by a smoother flow and simpler structure for a more fluent transition from positive to negative.
\subsection{Topic Control Task}
\begin{table*}[t]
\small
\centering
\begin{tabular}{clllcccc}
\toprule[1pt]
 &  & \multicolumn{2}{c}{\textbf{Control Strength(\%)}($\uparrow$)} & \textbf{Fluency}($\downarrow$ ) & \multicolumn{3}{c}{\textbf{Diverty}($\uparrow$)} \\ 
\multirow{-2}{*}{Topic} & \multirow{-2}{*}{Method} & Relevance & Correctness & PPL & Dist-1 & Dist-2 & Dist-3 \\ 
\midrule[1pt]
 & Fugde & 59.84 & 61.70 & 41.58 & 0.23 & 0.61 & 0.78 \\
 & DExpert & 76.56 & 77.90 & 41.44 & 0.20 & 0.65 & 0.84 \\
 & DisCup & 26.98 & 25.30 & 57.04 & 0.16 & 0.63 & 0.88 \\ 
\multirow{-4}{*}{World} & \ours & \textbf{84.14} & \textbf{84.20} & 35.92 & 0.17 & 0.60 & 0.83  \\ \midrule[0.5pt] 
 & Fugde & 60.67 & 61.10 & 39.81 & 0.20 & 0.63 & 0.81 \\
 & DExpert & 67.91 & 68.50 & 46.01 & 0.21 & 0.67 & 0.85 \\
 & DisCup & 35.49 & 35.90 & 60.80 & 0.15 & 0.60 & 0.88 \\ 
\multirow{-4}{*}{Sports} & \ours & {\textbf{97.49}} & \textbf{97.50} & 39.04 & 0.17 & 0.61 & 0.85 \\ \midrule[0.5pt] 
 & Fugde & 78.77 & 82.70 & 42.57 & 0.20 & 0.58 & 0.75 \\
 & DExpert & 45.60 & 46.40 & 42.21 & 0.22 & 0.66 & 0.83 \\
 & DisCup & 64.29 & 65.10 & 54.62 & 0.18 & 0.65 & 0.89 \\ 
\multirow{-4}{*}{Business} & \ours & {\textbf{88.28}} & {\textbf{88.60}} & 36.59 & 0.18 & 0.63 & 0.85 \\ \midrule[0.5pt] 
 & Fudge & 96.79 & \textbf{98.70} & 29.11 & 0.19 & 0.59 & 0.79 \\
 & DExpert & 96.46 &98.20 & 49.70 & 0.20 & 0.65 & 0.83 \\
 & DisCup & 87.80 & 91.50 & 58.29 & 0.17 & 0.63 & 0.88 \\ 
\multirow{-4}{*}{SciTech} & \ours & \textbf{97.21} & 97.70 & 44.23 & 0.19 & 0.64 & 0.86 \\ \midrule[0.5pt] 
 & Fugde & 74.02 $\pm 17.52$ & 76.05 $\pm 18.14$ & 38.27 & 0.21 & 0.60 & 0.78 \\
 & DExpert & 71.63 $\pm 21.07$ & 72.75 $\pm 21.50$ & 44.84 & 0.21 & 0.66 & 0.84 \\
 & DisCup & 53.64 $\pm 27.81$ & 54.45 $\pm 29.89$& 57.69 & 0.17 & 0.63 & 0.88 \\ 
\multirow{-4}{*}{Average} & \ours & \textbf{91.01 $\pm 7.74$}& \textbf{91.23} $\pm 7.87$& 40.57 & 0.18 & 0.62 & 0.85 \\ 
\bottomrule[1pt]
\vspace{1pt} \end{tabular}
\caption{Results on topic-controlled text generation. $\uparrow$ indicates that the higher the better, and $\downarrow$ is the opposite.}
\label{automatic_topic_result}
\end{table*}

\begin{figure}
\centering
		\includegraphics[width=0.85\linewidth]{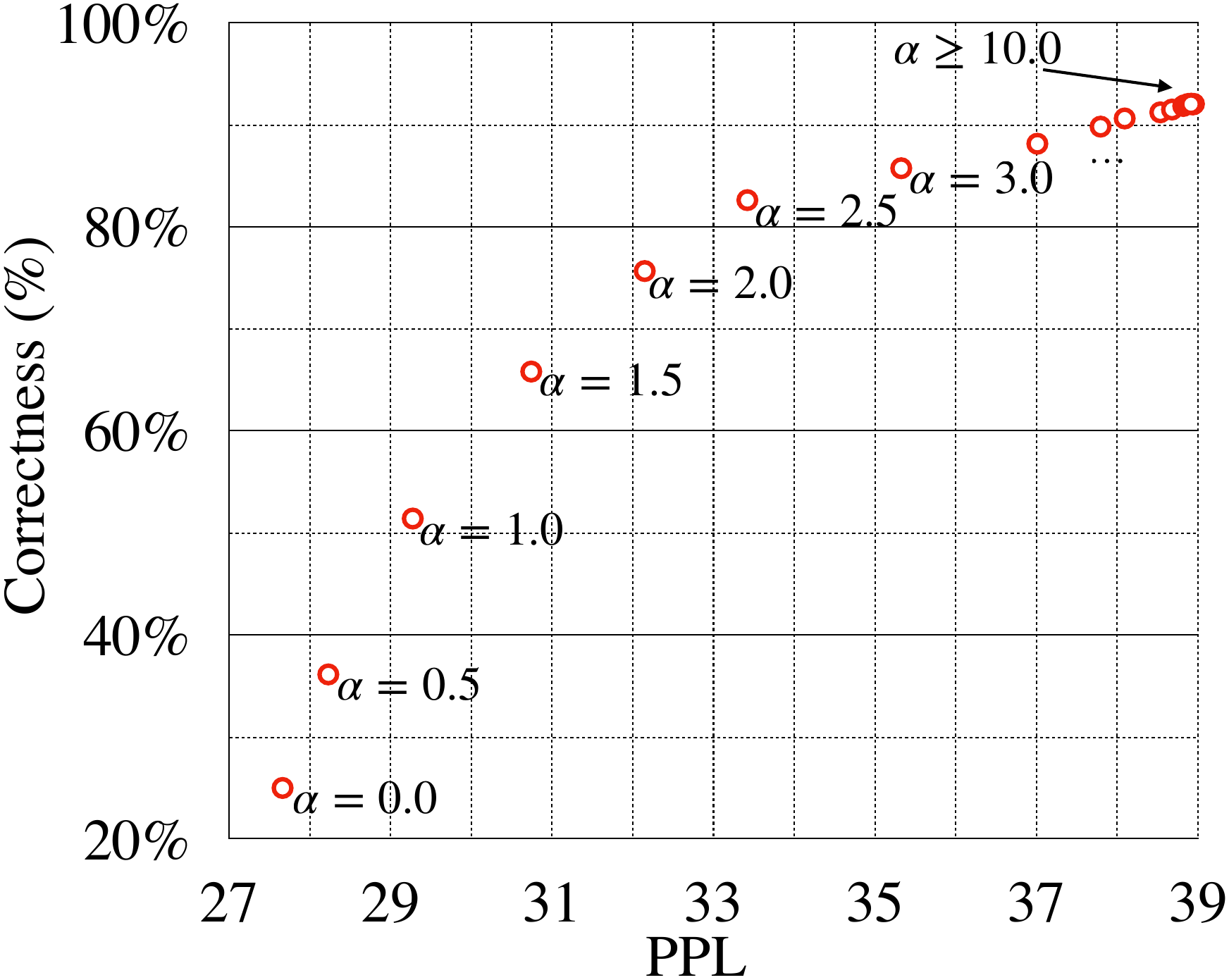}
		\caption{Effect of control strength (averaged on four topics).}
		\label{fig:alpha}
\end{figure}

\begin{table*}[]
\small
\centering
\begin{tabular}{p{0.08\textwidth}p{0.05\textwidth}p{0.80\textwidth}}
\toprule
\multicolumn{3}{c}{Prompt: In summary (\textbf{topic})} \\
\midrule
\multirow{6}{*}{World}& Fudge & In summary he was killed, apparently in an accident, in a police car with the police chief in it.If the two police cars were connected, it was a homicide, but of the five possible killers there were only three, all of \\
 & \ours & In summary of Sudan's Darfur operation, Khartoum says its troops will stop the fighting by 1 October, saying it had made all the necessary information available to the United Nations Security Council. The Sudanese military    says it will \\ \midrule[0.5pt] 
\multirow{6}{*}{Sports} & Fudge & In summary, the above is what the team decided.  All players with the option are on LT, all players with the option on RG and all players with the option on RT. If you notice a player, you won't be able to \\
 & \ours & In summary: The Patriots run out of time to score more points than Saints The NFL's worst team won't score again until the seventh quarter, which means the Saints will try to get in the end zone in an attempt to keep the Patriots from scoring \\ \midrule[0.5pt] 
\multirow{6}{*}{Business} & Fudge & In summary, an assessment shows a 10 percent to 36-percent excess and 5 to 8-percent adverse clinical or adverse drug adverse events in a sample of 514 patients with type 1 diabetes versus 486 who were given conventional therapy and no medications at \\ 
 & \ours & In summary, Google sees revenue per se down on ad revenue The Internet search and Web advertising business had its biggest decline during 2005 and 2006, according to financials offered by Google Inc. to investors in its initial September IPO, but a decline over \\ \midrule[0.5pt] 
\multirow{6}{*}{SciTech} & Fudge & In summary our goal in this blog to share about the many resources available to the beginner to explore and experiment with Arduino, C++, and Linux as the main Arduino or Arduino specific libraries that are supported.If you like this blog post and \\
 & \ours & In summary: Apple's iPod now compatible with Microsoft's Windows Media Player (WP) Windows Store Apple has confirmed this morning that Microsoft's long-awaited Windows XP Media Player 10 will allow for compatibility with the Microsoft Windows Update service, the iTunes Radio service
\\
\bottomrule
\vspace{1pt} \end{tabular}
\caption{Case study on topic control task.}
\label{case:topic}
\end{table*}
\begin{table}[t]
\small
\centering
\begin{tabular}{clcc}
\toprule[1pt]
\textbf{Task} & \textbf{Model} & \textbf{Win Rate}(\%) & \textbf{Fluency} \\
\midrule[1pt]
\multirow{2}{*}{Topic} & Fudge & 19 & 2.26 \\
 & \ours &  \textbf{62} & \textbf{2.77} \\ 
\bottomrule[1pt]
\vspace{1pt}  \end{tabular}
\caption{Human evaluation results on topic control task. Sign tests conducted on human scores demonstrate that \ours surpasses Fudge with a $p$-value of $<0.01$.}
\label{human_evaluation_result_exp_2}
\end{table}
Our second application is topic-controlled generation. The goal of topic control is to generate continuations that are on a given topic.
\paragraph{\textbf{Setup}}
Following previous works \citep{KrauseGedi2021, Qian2022ContrastivePrefixes}, we use the AGNews dataset \citep{Zhang2015AGNews}, which contains 4 topics (World, Sports, Business, and Sci/Tech) and 120K news articles in total. We use the first half of the original training data for training \ours and baselines, while the other half is used to train a RoBERTa \citep{Liu2019RoBERTa} classifier for measuring the control strength. 
We use the 20 prompts in PPLM \cite{Dathathri2020PPLM} for evaluation. For each prompt, 50 completions of length 50 are generated.
\paragraph{\textbf{Results}}
Table \ref{automatic_topic_result} presents the experimental results. On 3 out of 4 topics (World, Sports, and Business), our method achieves the highest \textit{Relevance} and \textit{Correctness} scores and the lowest PPLs. The only exception is on Sci/Tech, where Fudge has slightly higher \textit{Correctness} scores than \ours and a much lower PPL. However, as indicated by the diversity scores, the text generated by Fudge is less informative. We also report the average scores and standard deviations on control strength in the bottom block of Table \ref{automatic_topic_result}. On average, \ours outperforms the best baseline, Fudge, by 15.18 points (91.23\% vs. 76.05\%). In the meantime, \ours has the smallest standard deviation, which reveals the additional advantages of \ours: universality and robustness. In contrast, other baseline methods usually specialize in some topics and perform much worse in other topics. For example, DExpert is particularly good on SciTech but bad on Business (98.20\% vs. 46.40\% on \textit{Correctness}).
\paragraph{\textbf{Effect of Control Strength}}
We then turn to study the relationship between control strength and generated text. We run \ours using different values of $\alpha$ and plot the results in Figure \ref{fig:alpha}. As seen, the control strength increases as $\alpha$ gets larger. At the same time, PPL also increases, indicating higher control strength is more likely to hurt fluency. However, unlike weighted decoding methods where a large control strength causes unfluent sentences, \ours converges to a stable point. 

\paragraph{\textbf{Human Evaluation Result}}
Using the same experiment settings as detailed in the sentiment control task, we randomly sample 100 pairs of \ours and the best baseline model Fudge to ask for human evaluation.
The results stemming from the human evaluation are detailed in Table \ref{human_evaluation_result_exp_2}. Noteworthy is the substantial preference demonstrated by evaluators for \ours over Fudge in terms of control strength assessment for the topic control task, with a notable factor of 4.8. Moreover, in the aspect of fluency, \ours showcases a superior performance (+0.51) compared to Fudge. According to the significance tests, \ours showcases \ours exhibits substantial and statistically significant imprvements in terms of both control strength and fluency, as evidenced by $p$-values less than $0.01$.
\paragraph{\textbf{Case Study}} We select a subset of examples for an in-depth case study, as delineated in Table \ref{case:topic}. Take the prompt \textit{In summary for example}, for the World topic. \ours centered on Sudan's Darfur operation, holds higher relevance to global concerns due to its international implications and peacekeeping context. It presents information coherently, smoothly transitioning from the summary to the specifics of Khartoum's actions and engagement with the United Nations. This organized structure underscores the sentence's importance in ongoing World discussions. At the same time, the transition in the \ours is more natural. On the other hand, Fudge lacks explicit context and global significance. While it hints at a potential incident, it lacks clarity, leaving the reader uncertain about the event's broader impact. This absence of detail affects its relevance to world topics.
For the Sports topic, \ours is more relevant to sports topics due to its direct discussion of football teams, scoring, and game dynamics. Its fluency benefits from its coherent structure, use of sports terminology, and clear narrative. On the other hand, Fudge lacks context and explanation of technical terms, making it less relevant and potentially less clear.

\subsection{Stylistic Novel Writing}
\begin{table*}[t]
\small
\centering
\begin{tabular}{clllcccc}
\toprule[1pt]
 &  & \multicolumn{2}{c}{\textbf{Control Strength(\%)}($\uparrow$)} & \textbf{Fluency}($\downarrow$ ) & \multicolumn{3}{c}{\textbf{Diverty}($\uparrow$)} \\ 
\multirow{-2}{*}{Topic} & \multirow{-2}{*}{Method} & Relevance & Correctness & PPL & Dist-1 & Dist-2 & Dist-3 \\ 
\midrule[1pt]  
\multirow{4}{*}{Military} & Fudge & 57.18 & 56.88 & 24.08 & 0.03 & 0.33 & 0.63 \\
 & DExpert & 62.92 & 62.80 & 24.58 & 0.03 & 0.35 & 0.64 \\
 & DisCup & 69.80 & 69.68 & 77.76 & 0.03 & 0.39 & 0.75 \\ 
 & \ours & \textbf{73.75} & \textbf{73.60} & 23.92 & 0.02 & 0.27 & 0.56 \\ \midrule[0.5pt] 
\multirow{4}{*}{SCI-FI} & Fugde & 78.79 & 79.28 & 22.02 & 0.03 & 0.31 & 0.61 \\
 & DExpert & \textbf{88.50} & \textbf{88.56} & 22.39 & 0.03 & 0.30 & 0.61 \\
 & DisCup & 72.65 & 73.04 & 69.65 & 0.03 & 0.38 & 0.75 \\ 
 & \ours & 72.92 & 73.12 & 21.75 & 0.02 & 0.26 & 0.55 \\ \midrule[0.5pt] 
\multirow{4}{*}{Officialdom} & Fugde & 74.73 & 74.56 & 23.52 & 0.03 & 0.32 & 0.63 \\
 & DExpert & \textbf{84.68} & \textbf{85.04} & 23.20 & 0.03 & 0.33 & 0.62 \\
 & DisCup & 52.22 & 52.16 & 76.04 & 0.02 & 0.35 & 0.75 \\ 
 & \ours & 75.74 & 76.16 & 23.91 & 0.02 & 0.26 & 0.56 \\ \midrule[0.5pt] 
\multirow{4}{*}{MartialArt} & Fugde & 36.60 & 36.64 & 22.51 & 0.03 & 0.32 & 0.63 \\
 & DExpert & 52.33 & 52.56 & 22.23 & 0.03 & 0.32 & 0.63 \\
 & DisCup & 45.34 & 45.36 & 50.83 & 0.03 & 0.36 & 0.73 \\ 
 & \ours & \textbf{75.58} & \textbf{76.72} & 22.45 & 0.02 & 0.26 & 0.55 \\ \midrule[0.5pt] 
\multirow{4}{*}{Average} & Fudge & 61.83 $\pm 19.25$ & 61.84 $\pm 19.37$ & 23.03 & 0.03 & 0.32 & 0.63 \\
 & DExpert & 72.11 $\pm 17.34$ & 72.24 $\pm 17.38$ & 23.10 & 0.03 & 0.33 & 0.62 \\
 & DisCup & 60.01 $\pm 13.31$ & 60.06 $\pm 13.34$ & 68.57 & 0.03 & 0.37 & 0.74 \\ 
 & \ours & \textbf{74.50 $\pm 1.39$} & \textbf{74.90 $\pm 1.80$} & 23.01 & 0.02 & 0.26 & 0.55 \\ \bottomrule[1pt]
\vspace{1pt} \end{tabular}   
\caption{Results on style-controlled text generation. $\uparrow$ indicates that the higher the better, and $\downarrow$ is the opposite.}
\label{automatic_novel_result}
\end{table*}
\begin{table}
\small
\centering
\begin{tabular}{cccc}
\toprule[1pt]
factor $r$ & \#Param. & Correctness(\%) & PPL \\
\midrule[1pt]
16 &  1.04x &74.90 $\pm 1.80$ & \textbf{23.01} \\
8 &  1.09x &75.06 $\pm 1.45$ & 23.58 \\
4 & 1.17x & \textbf{77.04} $\pm 1.58$ & 23.80 \\
\toprule[1pt]
\vspace{1pt} \end{tabular}   
\caption{Comparison of different adapter sizes on style-controlled text generation. \# Param. is the relative size compared to the base LM.}
\label{automatic_adapter_size_result}
\end{table}

\begin{table}[t]
\small
\centering
\begin{tabular}{clcc}
\toprule[1pt]
\textbf{Task} & \textbf{Model} & \textbf{Win Rate}(\%) & \textbf{Fluency} \\
\midrule[1pt]
 \multirow{2}{*}{Novel} & DExperts & 16 & 2.26 \\
 & \ours & \textbf{69} & \textbf{2.54} \\
\bottomrule[1pt]
\vspace{1pt} \end{tabular}
\caption{Human evaluation results on stylistic novel writing. Sign tests conducted on human scores demonstrate that \ours surpasses DExperts with a $p$-value of $<0.01$.}
\label{human_evaluation_result_exp_3}
\end{table}
\begin{CJK}{UTF8}{gbsn}
\begin{table*}[]
\centering
\small
\begin{tabular}{p{0.08\textwidth}p{0.05\textwidth}p{0.8\textwidth}}
\toprule
\multicolumn{3}{c}{Prompt: 故事发生在麻省理工学院 Stories that happened at MIT} \\
\midrule
 &  & 故事发生在麻省理工学院附近一线。当地的法官当场以故意谋杀罪、抢劫罪嫌获刑9个月，并被捕处死3个月。据悉，本案涉及的三个美国保安官职位分别为： \\
 & \multirow{-2}{*}{DExpert} & The story takes place near the   Massachusetts Institute of Technology. The local judge sentenced him to 9   months in prison on the spot for intentional murder and robbery, and was   arrested and executed for 3 months. It is reported that the three U.S. security   officer positions involved in this case are: \\
 &  & 故事发生在麻省理工学院中。他们看着那些数千名德国人在学院中穿插在日军阵营中，就连一个德国老百姓也在听着这些德国同志的穿插言论。学生们听着这些 \\
\multirow{-7}{*}{Military} & \multirow{-2}{*}{\ours} & The story takes place at the   Massachusetts Institute of Technology. They watched those thousands of   Germans interspersed in the Japanese camp in the academy, and even a German   commoner was listening to the interspersed remarks of these German comrades. students listen to this \\ \midrule[0.5pt] 
 &  & 故事发生在麻省理工学院体验的那场真实版的恐怖游戏。这个游戏能够让更多人了解那个年代的日落。虽然有相似的情况，不过因为作者是年轻人，所以我们在 \\
 & \multirow{-2}{*}{DExpert} & The story takes place at the   base of the Massachusetts Institute of Technology. The protagonist of the   base, Al McIlroy, listened to the performance of the Disney band in the rain,   and Al McIlroy was able to enter the base. The work of the base was in the   maintenance room \\
 &  & 故事发生在麻省理工学院的一个实验室里，当时它就是一个空间传送实验室，由一个学生操纵其中一个人进行实验。通过这种方式他们才能发现一个不明飞行物 \\
\multirow{-7}{*}{SCI-FI} & \multirow{-2}{*}{\ours} & The story happened in a laboratory at the Massachusetts Institute of Technology, which was a space   transmission laboratory at the time, and a student manipulated one of them to   conduct experiments. This way they can spot a UFO \\ \midrule[0.5pt]
 &  & 故事发生在麻省理工学院，教学活动中一起大规模杀害的凶手被大面积破案。根据目击者的证词，他们被判刑十多年，并于2010年11月5日被释放。美国警方于1990 \\
 & \multirow{-2}{*}{DExpert} & The story happened at the   Massachusetts Institute of Technology, where a mass murderer in a teaching   activity was solved on a large scale. They were sentenced to more than ten   years in prison and released on November 5, 2010, based on eyewitness   testimony. US police in 1990 \\
 &  & 故事发生在麻省理工学院，是一所大学，学生的生活条件非常紧张，大家都在各自的办公室里休息，大家都在讨论着自己的一切工作，然后又是一场激烈的角力 \\
\multirow{-7}{*}{Officialdom} & \multirow{-2}{*}{\ours} & The story takes place in MIT,   a university where the living conditions of the students are very tense.   Everyone is resting in their respective offices. Everyone is discussing all   their work, and then there is a fierce wrestling \\ \midrule[0.5pt] 
 &  & 故事发生在麻省理工学院的体育馆，当时校园内聚集了不少高中的学院生，各个高中的毕业生也很多，有高中还有三年级的学生到这里为毕业晚会助威。这里的 \\
 & \multirow{-2}{*}{DExpert} & The story happened in the   gymnasium of the Massachusetts Institute of Technology. At that time, many   high school college students gathered on the campus, and there were also many   graduates from various high schools. Some high school and third-year students   came here to cheer for the graduation party. here \\
 &  & 故事发生在麻省理工学院东北约五分之二的位置，在此以前学校的官员们是全部来自美国，但是现在各不相同，有些人能出售一切东西，有些人就是来自于学校 \\
\multirow{-7}{*}{MartialArt} & \multirow{-2}{*}{\ours} & The story takes place about   two-fifths of the northeast of MIT. Before that, the school officials were   all from the United States, but now they are different. Some people can sell   everything, and some people are from the school. \\
\bottomrule
\vspace{1pt} \end{tabular}
\caption{Case study on stylistic novel writing.}
\label{case:novel1}
\end{table*}
\end{CJK}

Lastly, we introduce a new task, stylistic novel writing, for controlled text generation.
\paragraph{\textbf{Setup}}
We consider four novel styles including Military, SCI-FI, Officialdom, and MartialArt. For each style, we collect sentences from a Chinese novel website\footnote{\url{https://www.qidian.com/}}. The sentences are then presented to third-party annotators for annotation. In total, we collect 20K sentences for each style. Similar to the experiments in topic control, we use the first half data for training \ours and baselines. The second half is used for training a bert-base-chinese\footnote{bert-base-chinese} classifier for evaluating control strength. We also collect 80K random sentences without style labels. We choose 25 common story beginnings in Chinese as prompts (listed in Appendix \ref{appendix:novel}). For each of these prompts, 50 completions of length 70 are generated. We will release the dataset.

Unlike the previous experiments, we choose two Chinese GPT2 models \citep{Zhao2019uer} as our base LM and for training the discriminators in baselines respectively.\footnote{gpt2-chinese-cluecorpussmall and gpt2-distil-chinese-cluecorpussmall} We also use the base LM to test the perplexities of the generated continuations.
\paragraph{\textbf{Results}}
The results are shown in Table \ref{automatic_novel_result}. Overall, the situation is similar to that of topic control. On 2 out of 4 styles (Military and MartialArt), \ours method outperforms all baselines in terms of control strength and PPL. However, on SCI-FI and  Officialdom, \ours lags behind DExpert. Nevertheless, as shown in the last block of Table \ref{automatic_novel_result}, \ours is on average better than all baselines including DExpert.  The mixed result can be attributed to the large variances among different styles in the baselines. For example, the \textit{Correctness} scores of DExpert vary significantly between styles, with 88.56\% on SCI-FI and 52.56\% on Martial Art, a significant gap of 36 points. In contrast, \ours exhibits much more robustness, with standard deviations of \textit{Relevance} and \textit{Correctness} being 1.39\% and 1.80\% respectively.
\paragraph{\textbf{Effect of Adapter Size}}
Lastly, we study the effect of adapter size on performance. Recall that the size of adapters is controlled by a reduction factor. Throughout all previous experiments, the reduction factor is set to $16$. We show the results with larger adapters (smaller reduction factors) in Table \ref{automatic_adapter_size_result}. As can be seen, as the adapter size increases, we can achieve stronger control strength. However, the PPL also increases, likely due to the model being able to deviate further from the base LM.

\paragraph{\textbf{Human Evaluation Result}}
In the domain of the stylistic novel writing task, we conduct a comprehensive comparison between our model (\ours) and the prominent baseline model, DExperts, similar to the sentiment control and topic control tasks. We subsequently detail the outcomes of the human evaluation in Table \ref{human_evaluation_result_exp_3}. Particularly notable is the substantial preference expressed by evaluators for \ours over the baseline models, with a notable factor of 4.3, in terms of control strength assessment for the stylistic novel writing. Furthermore, in terms of fluency, \ours exhibits superior performance (+0.28) compared to the highly robust baseline model, DExperts. Based on the results of the significance tests, \ours showcases notable and statistically significant advantages in terms of both control strength and fluency, as supported by a $p$-value of less than $0.01$.
\paragraph{\textbf{Case Study}}
We select a subset of examples for an in-depth case study, as outlined in Table \ref{case:novel1}. Given that Stylistic Novel Writing involves crafting text in Chinese, we translate it into English for enhanced presentation. Note that this task poses unique challenges due to the nuanced nature of attributes, making their definition complex. In the context of Military writing, \ours stands out by explicitly situating the story at the Massachusetts Institute of Technology and depicting scenes of Germans infiltrating a Japanese military camp on campus. These settings and scenarios are intricately tied to the military theme, effectively conveying elements of a military backdrop. Conversely, while DExpert references judges and crimes, it does not directly delve into military themes. Furthermore, \ours provides comprehensive and vivid descriptions, smoothly transitioning between sentences.
For narratives with a SCI-FI theme, \ours once again aligns seamlessly with the subject matter. It portrays a space teleportation experiment in a laboratory involving unidentified flying objects, closely intertwining with science fiction elements. This sentence adeptly captures the essence of the science fiction plot, presenting it coherently and fluidly. In contrast, DExpert touches upon horror games and an era's conclusion, which are less directly linked to the science fiction theme. Furthermore, this sentence experiences structural disruptions and lacks a complete expression.

\section{Conclusion}
We presented \ours, a lightweight method for controlled text generation that uses fine-grained control codes and parameter-efficient adapters. The experiments on sentiment control, topic control, and style control show\sout{ed} that \ours consistently achieves strong performance on steering an off-the-shelf language model towards the desired attributes. Additionally, \ours only introduces a small number of new parameters to the base LM.

\appendix
\label{appendix:human}

\section{Test Prompts for Stylistic Novel Writing}
\label{appendix:novel}

We present the 25 Chinese prompts used for the stylistic novel writing task. These prompts cover a variety of themes, emotions, and settings that serve as a comprehensive testbed for our model. The complete set of prompts and their corresponding English translations are provided in Table \ref{novel_prompts}.
These prompts serve as a basis for evaluating our model's ability to generate stylistic novel content, allowing us to assess its performance and compare it with other baseline models. By using a diverse set of prompts, we can better understand the strengths and limitations of our model in generating creative and coherent narratives.

\begin{CJK}{UTF8}{gbsn}
\begin{table*}[t]
\small
\centering
\begin{tabular}{p{0.3\textwidth}p{0.6\textwidth}}
\toprule
Chinese Prompt & English translation \\
\midrule
这是一个动人的故事 & This is a moving story \\
这是一个悲伤的故事 & This is a sad story \\
这是一个励志的故事 & This is an inspirational story \\
此事还得从这说起 & This matter has to start from here \\
这是杨过第一次遇到郭靖 & This is the first time Yang Guo met Guo Jing \\
这是杜拉拉第一次遇到罗杰 & This is the first time Du Lala met Roger \\
这是叶文洁第一次遇到史强 & This is the first time Ye Wenjie met Shi Qiang \\
这是李云龙第一次遇到楚云飞 & This is the first time Li Yunlong met Chu Yunfei \\
那是一个春天 & It was a spring \\
那是一个夏天 & It was a summer \\
那是一个秋天 & It was an autumn \\
那是一个冬天 & It was a winter \\
故事发生在麻省理工学院 & Stories that happened at MIT \\
故事发生在东北大地 & The story takes place in the Northeast \\
故事发生在华山之巅 & The story happened on the top of Mount Hua \\
故事发生在上海机场 & The story happened at the Shanghai airport \\
谁也无法预知一个名不经传的小人物未来是否能够名震天下 & No one can predict whether a little-known person will become   famous all over the world in the future \\
黄昏的荒原远方悬着一轮红日 & A red sun hangs in the distance in the wasteland at dusk \\
屋内坐着一位头发花白的老妇人 & In the house sat an old woman with gray hair \\
我终于回到了阔别数年的故乡 & I finally returned to my hometown where I had been away for   several years \\
你永远不知道下一秒会发生什么 & You never know what will happen next second \\
一个人提着箱子进了长安城 & A man entered Chang'an City with a suitcase \\
屋里坐着一位穿着怪异的男子 & In the room sat a strangely dressed man \\
这是一个晴朗的早晨 & It's a sunny morning \\
这是一个安静的晚上 & It's a quiet night \\
\bottomrule[1pt]
\vspace{1pt} \end{tabular}
\caption{Test prompts for stylistic novel writing.}
\label{novel_prompts}
\end{table*}
\end{CJK}

\bibliographystyle{IEEEbib}
\bibliography{ref}

\begin{thebibliography}{10}

\bibitem{radford2019language}
A.~Radford, J.~Wu, R.~Child, D.~Luan, D.~Amodei, I.~Sutskever, et~al.,
\newblock ``Language models are unsupervised multitask learners,''
\newblock {\em OpenAI blog}, vol. 1, no. 8, pp. 9, 2019.

\bibitem{brown2020language}
T.~Brown, B.~Mann, N.~Ryder, M.~Subbiah, J.~D. Kaplan, P.~Dhariwal,
  A.~Neelakantan, P.~Shyam, G.~Sastry, A.~Askell, et~al.,
\newblock ``Language models are few-shot learners,''
\newblock {\em Advances in neural information processing systems}, vol. 33, pp.
  1877--1901, 2020.

\bibitem{RaffelSRLNMZLL20}
C.~Raffel, N.~Shazeer, A.~Roberts, K.~Lee, S.~Narang, M.~Matena, Y.~Zhou,
  W.~Li, and P.~J. Liu,
\newblock ``Exploring the limits of transfer learning with a unified
  text-to-text transformer,''
\newblock {\em J. Mach. Learn. Res.}, vol. 21, pp. 140:1--140:67, 2020.

\bibitem{Ficler2017Finetune}
J.~Ficler and Y.~Goldberg,
\newblock ``Controlling linguistic style aspects in neural language
  generation,''
\newblock {\em CoRR}, vol. abs/1707.02633, 2017.

\bibitem{Yu2017Finetune}
L.~Yu, W.~Zhang, J.~Wang, and Y.~Yu,
\newblock ``Seqgan: Sequence generative adversarial nets with policy
  gradient,''
\newblock in {\em Proceedings of the Thirty-First {AAAI} Conference on
  Artificial Intelligence, February 4-9, 2017, San Francisco, California,
  {USA}}, S.~Singh and S.~Markovitch, Eds. 2017, pp. 2852--2858, {AAAI} Press.

\bibitem{Ziegler2019Finetune}
D.~M. Ziegler, N.~Stiennon, J.~Wu, T.~B. Brown, A.~Radford, D.~Amodei, P.~F.
  Christiano, and G.~Irving,
\newblock ``Fine-tuning language models from human preferences,''
\newblock {\em CoRR}, vol. abs/1909.08593, 2019.

\bibitem{Dathathri2020PPLM}
S.~Dathathri, A.~Madotto, J.~Lan, J.~Hung, E.~Frank, P.~Molino, J.~Yosinski,
  and R.~Liu,
\newblock ``Plug and play language models: {A} simple approach to controlled
  text generation,''
\newblock in {\em 8th International Conference on Learning Representations,
  {ICLR} 2020, Addis Ababa, Ethiopia, April 26-30, 2020}. 2020, OpenReview.net.

\bibitem{KrauseGedi2021}
B.~Krause, A.~D. Gotmare, B.~McCann, N.~S. Keskar, S.~R. Joty, R.~Socher, and
  N.~F. Rajani,
\newblock ``Gedi: Generative discriminator guided sequence generation,''
\newblock in {\em Findings of the Association for Computational Linguistics:
  {EMNLP} 2021, Virtual Event / Punta Cana, Dominican Republic, 16-20 November,
  2021}, M.~Moens, X.~Huang, L.~Specia, and S.~W. Yih, Eds. 2021, pp.
  4929--4952, Association for Computational Linguistics.

\bibitem{Yang2021FDUGE}
K.~Yang and D.~Klein,
\newblock ``{FUDGE:} controlled text generation with future discriminators,''
\newblock in {\em Proceedings of the 2021 Conference of the North American
  Chapter of the Association for Computational Linguistics: Human Language
  Technologies, {NAACL-HLT} 2021, Online, June 6-11, 2021}, K.~Toutanova,
  A.~Rumshisky, L.~Zettlemoyer, D.~Hakkani{-}T{\"{u}}r, I.~Beltagy, S.~Bethard,
  R.~Cotterell, T.~Chakraborty, and Y.~Zhou, Eds. 2021, pp. 3511--3535,
  Association for Computational Linguistics.

\bibitem{Liu2021DExperts}
A.~Liu, M.~Sap, X.~Lu, S.~Swayamdipta, C.~Bhagavatula, N.~A. Smith, and
  Y.~Choi,
\newblock ``Dexperts: Decoding-time controlled text generation with experts and
  anti-experts,''
\newblock in {\em Proceedings of the 59th Annual Meeting of the Association for
  Computational Linguistics and the 11th International Joint Conference on
  Natural Language Processing, {ACL/IJCNLP} 2021, (Volume 1: Long Papers),
  Virtual Event, August 1-6, 2021}, C.~Zong, F.~Xia, W.~Li, and R.~Navigli,
  Eds. 2021, pp. 6691--6706, Association for Computational Linguistics.

\bibitem{Arora2022Director}
K.~Arora, K.~Shuster, S.~Sukhbaatar, and J.~Weston,
\newblock ``Director: Generator-classifiers for supervised language modeling,''
\newblock in {\em Proceedings of the 2nd Conference of the Asia-Pacific Chapter
  of the Association for Computational Linguistics and the 12th International
  Joint Conference on Natural Language Processing, {AACL/IJCNLP} 2022 - Volume
  1: Long Papers, Online Only, November 20-23, 2022}, Y.~He, H.~Ji, Y.~Liu,
  S.~Li, C.~Chang, S.~Poria, C.~Lin, W.~L. Buntine, M.~Liakata, H.~Yan, Z.~Yan,
  S.~Ruder, X.~Wan, M.~Arana{-}Catania, Z.~Wei, H.~Huang, J.~Wu, M.~Day,
  P.~Liu, and R.~Xu, Eds. 2022, pp. 512--526, Association for Computational
  Linguistics.

\bibitem{Qian2022ContrastivePrefixes}
J.~Qian, L.~Dong, Y.~Shen, F.~Wei, and W.~Chen,
\newblock ``Controllable natural language generation with contrastive
  prefixes,''
\newblock in {\em Findings of the Association for Computational Linguistics:
  {ACL} 2022, Dublin, Ireland, May 22-27, 2022}, S.~Muresan, P.~Nakov, and
  A.~Villavicencio, Eds. 2022, pp. 2912--2924, Association for Computational
  Linguistics.

\bibitem{Zhang2022DisCup}
H.~Zhang and D.~Song,
\newblock ``Discup: Discriminator cooperative unlikelihood prompt-tuning for
  controllable text generation,''
\newblock {\em CoRR}, vol. abs/2210.09551, 2022.

\bibitem{Yang2022Tailor}
K.~Yang, D.~Liu, W.~Lei, B.~Yang, M.~Xue, B.~Chen, and J.~Xie,
\newblock ``Tailor: {A} prompt-based approach to attribute-based controlled
  text generation,''
\newblock {\em CoRR}, vol. abs/2204.13362, 2022.

\bibitem{GU2022DistributionalLens}
Y.~Gu, X.~Feng, S.~Ma, L.~Zhang, H.~Gong, and B.~Qin,
\newblock ``A distributional lens for multi-aspect controllable text
  generation,''
\newblock {\em CoRR}, vol. abs/2210.02889, 2022.

\bibitem{GuFMZGZ023}
Y.~Gu, X.~Feng, S.~Ma, L.~Zhang, H.~Gong, W.~Zhong, and B.~Qin,
\newblock ``Controllable text generation via probability density estimation in
  the latent space,''
\newblock in {\em Proceedings of the 61st Annual Meeting of the Association for
  Computational Linguistics (Volume 1: Long Papers), {ACL} 2023, Toronto,
  Canada, July 9-14, 2023}, A.~Rogers, J.~L. Boyd{-}Graber, and N.~Okazaki,
  Eds. 2023, pp. 12590--12616, Association for Computational Linguistics.

\bibitem{houlsby2019parameter}
N.~Houlsby, A.~Giurgiu, S.~Jastrzebski, B.~Morrone, Q.~De~Laroussilhe,
  A.~Gesmundo, M.~Attariyan, and S.~Gelly,
\newblock ``Parameter-efficient transfer learning for nlp,''
\newblock in {\em International Conference on Machine Learning}. PMLR, 2019,
  pp. 2790--2799.

\bibitem{pfeiffer2021adapterfusion}
J.~Pfeiffer, A.~Kamath, A.~R{\"u}ckl{\'e}, K.~Cho, and I.~Gurevych,
\newblock ``Adapterfusion: Non-destructive task composition for transfer
  learning,''
\newblock in {\em Proceedings of the 16th Conference of the European Chapter of
  the Association for Computational Linguistics: Main Volume}, 2021, pp.
  487--503.

\bibitem{he2021towards}
J.~He, C.~Zhou, X.~Ma, T.~Berg-Kirkpatrick, and G.~Neubig,
\newblock ``Towards a unified view of parameter-efficient transfer learning,''
\newblock in {\em International Conference on Learning Representations}, 2021.

\bibitem{Keskar2019CTRL}
N.~S. Keskar, B.~McCann, L.~R. Varshney, C.~Xiong, and R.~Socher,
\newblock ``{CTRL:} {A} conditional transformer language model for controllable
  generation,''
\newblock {\em CoRR}, vol. abs/1909.05858, 2019.

\bibitem{Chan2021CoCon}
A.~Chan, Y.~Ong, B.~Pung, A.~Zhang, and J.~Fu,
\newblock ``Cocon: {A} self-supervised approach for controlled text
  generation,''
\newblock in {\em 9th International Conference on Learning Representations,
  {ICLR} 2021, Virtual Event, Austria, May 3-7, 2021}. 2021, OpenReview.net.

\bibitem{Yu2021AttributeAlignment}
D.~Yu, Z.~Yu, and K.~Sagae,
\newblock ``Attribute alignment: Controlling text generation from pre-trained
  language models,''
\newblock in {\em Findings of the Association for Computational Linguistics:
  {EMNLP} 2021, Virtual Event / Punta Cana, Dominican Republic, 16-20 November,
  2021}, M.~Moens, X.~Huang, L.~Specia, and S.~W. Yih, Eds. 2021, pp.
  2251--2268, Association for Computational Linguistics.

\bibitem{carlsson-etal-2022-fine}
F.~Carlsson, J.~{\"O}hman, F.~Liu, S.~Verlinden, J.~Nivre, and M.~Sahlgren,
\newblock ``Fine-grained controllable text generation using non-residual
  prompting,''
\newblock in {\em Proceedings of the 60th Annual Meeting of the Association for
  Computational Linguistics (Volume 1: Long Papers)}, Dublin, Ireland, May
  2022, pp. 6837--6857, Association for Computational Linguistics.

\bibitem{Ghazvininejad2017WD}
M.~Ghazvininejad, X.~Shi, J.~Priyadarshi, and K.~Knight,
\newblock ``Hafez: an interactive poetry generation system,''
\newblock in {\em Proceedings of the 55th Annual Meeting of the Association for
  Computational Linguistics, {ACL} 2017, Vancouver, Canada, July 30 - August 4,
  System Demonstrations}, M.~Bansal and H.~Ji, Eds. 2017, pp. 43--48,
  Association for Computational Linguistics.

\bibitem{Holtzman2018WD}
A.~Holtzman, J.~Buys, M.~Forbes, A.~Bosselut, D.~Golub, and Y.~Choi,
\newblock ``Learning to write with cooperative discriminators,''
\newblock in {\em Proceedings of the 56th Annual Meeting of the Association for
  Computational Linguistics, {ACL} 2018, Melbourne, Australia, July 15-20,
  2018, Volume 1: Long Papers}, I.~Gurevych and Y.~Miyao, Eds. 2018, pp.
  1638--1649, Association for Computational Linguistics.

\bibitem{Gordon2018WD}
R.~Cohn{-}Gordon, N.~D. Goodman, and C.~Potts,
\newblock ``Pragmatically informative image captioning with character-level
  inference,''
\newblock in {\em Proceedings of the 2018 Conference of the North American
  Chapter of the Association for Computational Linguistics: Human Language
  Technologies, NAACL-HLT, New Orleans, Louisiana, USA, June 1-6, 2018, Volume
  2 (Short Papers)}, M.~A. Walker, H.~Ji, and A.~Stent, Eds. 2018, pp.
  439--443, Association for Computational Linguistics.

\bibitem{Shen2019WD}
S.~Shen, D.~Fried, J.~Andreas, and D.~Klein,
\newblock ``Pragmatically informative text generation,''
\newblock in {\em Proceedings of the 2019 Conference of the North American
  Chapter of the Association for Computational Linguistics: Human Language
  Technologies, {NAACL-HLT} 2019, Minneapolis, MN, USA, June 2-7, 2019, Volume
  1 (Long and Short Papers)}, J.~Burstein, C.~Doran, and T.~Solorio, Eds. 2019,
  pp. 4060--4067, Association for Computational Linguistics.

\bibitem{GU2022CATPAW}
Y.~Gu, X.~Feng, S.~Ma, J.~Wu, H.~Gong, and B.~Qin,
\newblock ``Improving controllable text generation with position-aware weighted
  decoding,''
\newblock in {\em Findings of the Association for Computational Linguistics:
  {ACL} 2022, Dublin, Ireland, May 22-27, 2022}, S.~Muresan, P.~Nakov, and
  A.~Villavicencio, Eds. 2022, pp. 3449--3467, Association for Computational
  Linguistics.

\bibitem{Mireshghallah2022MixandMatch}
F.~Mireshghallah, K.~Goyal, and T.~Berg{-}Kirkpatrick,
\newblock ``Mix and match: Learning-free controllable text generation using
  energy language models,''
\newblock {\em CoRR}, vol. abs/2203.13299, 2022.

\bibitem{Yann2006EnergyBased}
Y.~LeCun, S.~Chopra, R.~Hadsell, M.~Ranzato, and F.~Huang,
\newblock ``A tutorial on energy-based learning,''
\newblock {\em Predicting structured data}, vol. 1, no. 0, 2006.

\bibitem{Li2021PrefixTuning}
X.~L. Li and P.~Liang,
\newblock ``Prefix-tuning: Optimizing continuous prompts for generation,''
\newblock in {\em Proceedings of the 59th Annual Meeting of the Association for
  Computational Linguistics and the 11th International Joint Conference on
  Natural Language Processing, {ACL/IJCNLP} 2021, (Volume 1: Long Papers),
  Virtual Event, August 1-6, 2021}, C.~Zong, F.~Xia, W.~Li, and R.~Navigli,
  Eds. 2021, pp. 4582--4597, Association for Computational Linguistics.

\bibitem{vaswani2017attention}
A.~Vaswani, N.~Shazeer, N.~Parmar, J.~Uszkoreit, L.~Jones, A.~N. Gomez,
  {\L}.~Kaiser, and I.~Polosukhin,
\newblock ``Attention is all you need,''
\newblock {\em Advances in neural information processing systems}, vol. 30,
  2017.

\bibitem{he2016deep}
K.~He, X.~Zhang, S.~Ren, and J.~Sun,
\newblock ``Deep residual learning for image recognition,''
\newblock in {\em Proceedings of the IEEE conference on computer vision and
  pattern recognition}, 2016, pp. 770--778.

\bibitem{ba2016layer}
J.~L. Ba, J.~R. Kiros, and G.~E. Hinton,
\newblock ``Layer normalization,''
\newblock {\em arXiv preprint arXiv:1607.06450}, 2016.

\bibitem{liu-etal-2019-linguistic}
N.~F. Liu, M.~Gardner, Y.~Belinkov, M.~E. Peters, and N.~A. Smith,
\newblock ``Linguistic knowledge and transferability of contextual
  representations,''
\newblock in {\em Proceedings of the 2019 Conference of the North {A}merican
  Chapter of the Association for Computational Linguistics: Human Language
  Technologies, Volume 1 (Long and Short Papers)}, Minneapolis, Minnesota, June
  2019, pp. 1073--1094, Association for Computational Linguistics.

\bibitem{Li2016Dist}
J.~Li, M.~Galley, C.~Brockett, J.~Gao, and B.~Dolan,
\newblock ``A diversity-promoting objective function for neural conversation
  models,''
\newblock in {\em {NAACL} {HLT} 2016, The 2016 Conference of the North American
  Chapter of the Association for Computational Linguistics: Human Language
  Technologies, San Diego California, USA, June 12-17, 2016}, K.~Knight,
  A.~Nenkova, and O.~Rambow, Eds. 2016, pp. 110--119, The Association for
  Computational Linguistics.

\bibitem{Li2018Sentiment}
J.~Li, R.~Jia, H.~He, and P.~Liang,
\newblock ``Delete, retrieve, generate: a simple approach to sentiment and
  style transfer,''
\newblock in {\em Proceedings of the 2018 Conference of the North {A}merican
  Chapter of the Association for Computational Linguistics: Human Language
  Technologies, Volume 1 (Long Papers)}, New Orleans, Louisiana, June 2018, pp.
  1865--1874, Association for Computational Linguistics.

\bibitem{Sudhakar2019Sentiment}
A.~Sudhakar, B.~Upadhyay, and A.~Maheswaran,
\newblock ``{``}transforming{''} delete, retrieve, generate approach for
  controlled text style transfer,''
\newblock in {\em Proceedings of the 2019 Conference on Empirical Methods in
  Natural Language Processing and the 9th International Joint Conference on
  Natural Language Processing (EMNLP-IJCNLP)}, Hong Kong, China, Nov. 2019, pp.
  3269--3279, Association for Computational Linguistics.

\bibitem{Socher2013SST5}
R.~Socher, A.~Perelygin, J.~Wu, J.~Chuang, C.~D. Manning, A.~Y. Ng, and
  C.~Potts,
\newblock ``Recursive deep models for semantic compositionality over a
  sentiment treebank,''
\newblock in {\em Proceedings of the 2013 Conference on Empirical Methods in
  Natural Language Processing, {EMNLP} 2013, 18-21 October 2013, Grand Hyatt
  Seattle, Seattle, Washington, USA, {A} meeting of SIGDAT, a Special Interest
  Group of the {ACL}}. 2013, pp. 1631--1642, {ACL}.

\bibitem{Gokaslan2019OpenWebText}
A.~Gokaslan and V.~Cohen,
\newblock ``Openwebtext corpus,'' 2019.

\bibitem{maas-etal-2011-learning}
A.~L. Maas, R.~E. Daly, P.~T. Pham, D.~Huang, A.~Y. Ng, and C.~Potts,
\newblock ``Learning word vectors for sentiment analysis,''
\newblock in {\em Proceedings of the 49th Annual Meeting of the Association for
  Computational Linguistics: Human Language Technologies}, Portland, Oregon,
  USA, June 2011, pp. 142--150, Association for Computational Linguistics.

\bibitem{Zhang2015AGNews}
X.~Zhang, J.~J. Zhao, and Y.~LeCun,
\newblock ``Character-level convolutional networks for text classification,''
\newblock in {\em Advances in Neural Information Processing Systems 28: Annual
  Conference on Neural Information Processing Systems 2015, December 7-12,
  2015, Montreal, Quebec, Canada}, C.~Cortes, N.~D. Lawrence, D.~D. Lee,
  M.~Sugiyama, and R.~Garnett, Eds., 2015, pp. 649--657.

\bibitem{Liu2019RoBERTa}
Y.~Liu, M.~Ott, N.~Goyal, J.~Du, M.~Joshi, D.~Chen, O.~Levy, M.~Lewis,
  L.~Zettlemoyer, and V.~Stoyanov,
\newblock ``Roberta: {A} robustly optimized {BERT} pretraining approach,''
\newblock {\em CoRR}, vol. abs/1907.11692, 2019.

\bibitem{Zhao2019uer}
Z.~Zhao, H.~Chen, J.~Zhang, X.~Zhao, T.~Liu, W.~Lu, X.~Chen, H.~Deng, Q.~Ju,
  and X.~Du,
\newblock ``Uer: An open-source toolkit for pre-training models,''
\newblock {\em EMNLP-IJCNLP 2019}, p. 241, 2019.

\end{thebibliography}

\vfill

\end{document}